\documentclass[conference,compsoc]{IEEEtran}

\usepackage{graphicx}
\usepackage{lipsum}
\usepackage{multirow}
\usepackage{booktabs}
\usepackage{amssymb}
\usepackage{amsmath}
\usepackage[margin=1in]{geometry}
\usepackage[pagebackref,breaklinks,colorlinks]{hyperref}

\ifCLASSOPTIONcompsoc
  \usepackage[nocompress]{cite}
\else
  \usepackage{cite}
\fi
\ifCLASSINFOpdf
\else
\fi
\begin{document}
%
\title{Video Forgery Detection with Optical Flow Residuals and Spatial-Temporal Consistency}


\author{
Xi Xue, Kunio Suzuki, Nabarun Goswami and Takuya Shintate \\
NABLAS Inc., Tokyo, Japan
}

\maketitle

\begin{abstract}
The rapid advancement of diffusion-based video generation models has led to increasingly realistic synthetic content, presenting new challenges for video forgery detection. Existing methods often struggle to capture fine-grained temporal inconsistencies, particularly in AI-generated videos with high visual fidelity and coherent motion. In this work, we propose a detection framework that leverages spatial-temporal consistency by combining RGB appearance features with optical flow residuals. The model adopts a dual-branch architecture, where one branch analyzes RGB frames to detect appearance-level artifacts, while the other processes flow residuals to reveal subtle motion anomalies caused by imperfect temporal synthesis. By integrating these complementary features, the proposed method effectively detects a wide range of forged videos. Extensive experiments on text-to-video and image-to-video tasks across ten diverse generative models demonstrate the robustness and strong generalization ability of the proposed approach.
\end{abstract}


%
\IEEEpeerreviewmaketitle

\section{Introduction}

AI-generated content (AIGC) has gained increasing popularity in recent years, spanning diverse domains such as computer vision, natural language processing, and speech synthesis. In the field of computer vision, significant advancements in generative models, particularly the introduction of Stable Diffusion \cite{stable}, have substantially improved the realism and controllability of synthetic images. Building on this momentum, research has gradually shifted from static image generation to video synthesis, with growing emphasis on generating temporally coherent and visually consistent sequences.

While these models open new opportunities for creativity and media production, they also pose significant risks, particularly through the spread of misinformation and the creation of deliberately manipulated content. As a result, video forgery detection has become a critical task in multimedia forensics and content authentication.

Early work on video forgery detection mainly focused on facial deepfakes, such as face swapping \cite{swap1, swap2} and lip-syncing \cite{lip1, lip2}. These approaches typically rely on detecting spatial-level artifacts, including unnatural facial textures, blending inconsistencies, or frequency-domain anomalies. While effective in constrained scenarios, such methods generalize poorly to more diverse synthetic content.

With the rise of advanced diffusion-based video generation models such as Pika \cite{pika}, Sora \cite{sora}, and VideoCrafter \cite{videocraft}, synthetic content has expanded beyond facial videos to cover a wide range of scenes, objects, and motion patterns. This shift presents new challenges, as many traditional detectors struggle with forged content that lacks obvious visual artifacts.

In AI-generated video forgery detection, most existing methods focus on spatial appearance or temporal consistency. For temporal modeling, prior studies have investigated LSTM-based recurrent networks \cite{turns} and optical flow representations \cite{distinguish, aigvdet, matters}. Although LSTMs theoretically support long-term dependency modeling, they often suffer from vanishing gradients and limited memory in practice. Meanwhile, optical flow maps primarily encode global motion, making them less effective at exposing localized temporal inconsistencies.

To address these limitations, we propose a dual-branch detection framework that jointly models spatial and temporal anomalies. The spatial branch analyzes RGB frames to detect appearance-level artifacts such as texture distortions and unnatural object boundaries. The temporal branch leverages optical flow residuals, computed as differences between consecutive flow maps, to amplify localized motion anomalies while suppressing global motion. This design enables the model to detect temporally implausible patterns that may appear visually coherent but are physically inconsistent.

We evaluate our approach on two representative generation tasks, text-to-video (T2V) and image-to-video (I2V), using ten datasets produced by diverse diffusion-based models. Our proposed method achieves state-of-the-art performance across all benchmarks, consistently delivering high accuracy, AUC, and F1 scores. These results confirm the robustness and generalization capability of our method in detecting a wide range of AI-generated forgeries.

The remainder of this paper is organized as follows. Section 2 reviews related work on video generation and forgery detection. Section 3 details the proposed detection framework. Section 4 presents the experimental setup and results. Finally, Section 5 concludes the paper.

\begin{figure*}[!t]
\centering
\includegraphics[width=0.95\linewidth]{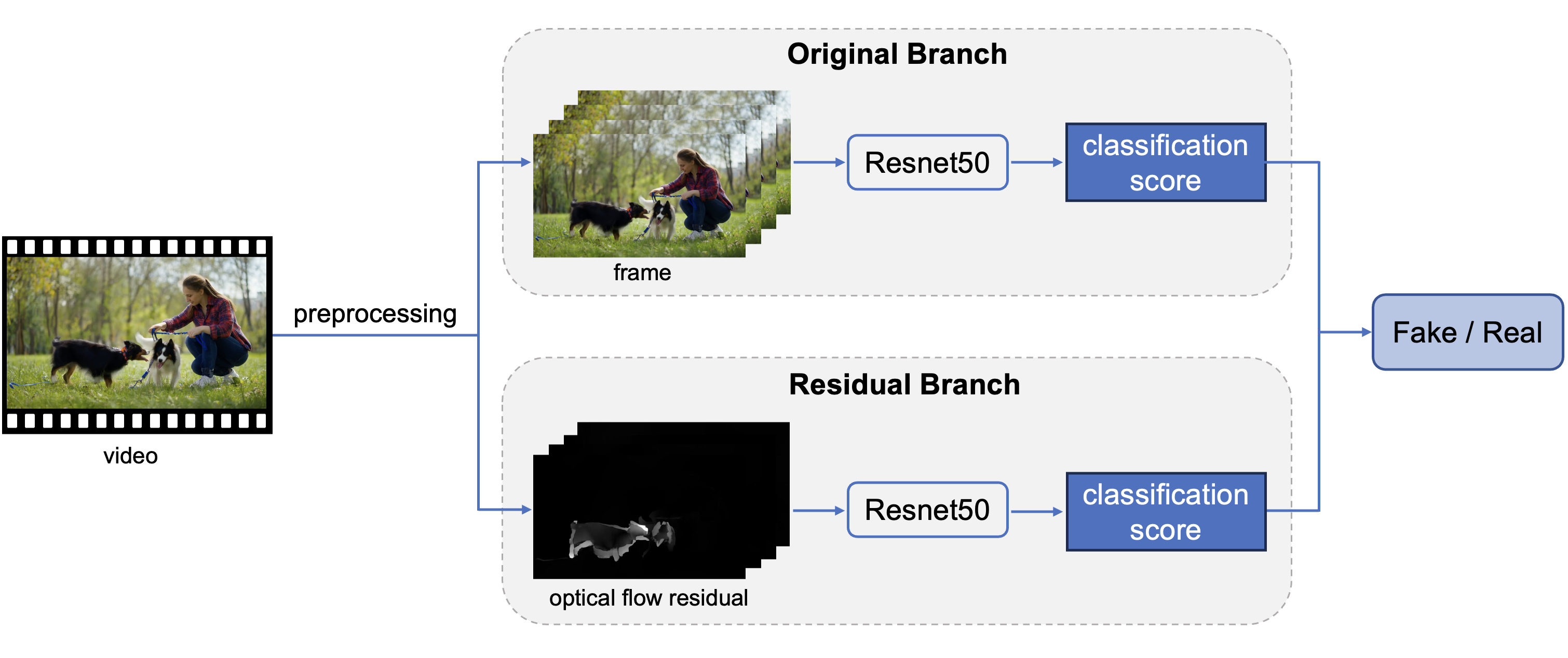}
\caption{\label{fig:architecture}Overview of the proposed architecture.}
\end{figure*}

\section{Related Work}
\subsection{Video Diffusion Models}

Early diffusion-based video generation was pioneered by models such as Imagen Video \cite{imagen} and Make-A-Video \cite{make-a-video}, which adopted frame-wise or cascaded generation strategies. Although some temporal smoothing or sparse temporal attention was applied, these methods often suffered from motion flickering and inconsistencies due to insufficient spatio-temporal coherence.

To address these limitations, recent works have introduced Diffusion Transformers (DiT) \cite{dit}, which enhance spatio-temporal modeling by incorporating space-time attention and 3D latent denoising within the generative process.
Notable models such as Pika \cite{pika}, AnimateDiff \cite{animatediff}, and Sora \cite{sora} adopt DiT-based architectures for tasks including text-to-video generation and motion-guided animation. These models demonstrate significantly improved temporal coherence and visual realism, marking a substantial step forward in diffusion-based video synthesis.

These advances have significantly increased the realism and diversity of AI-generated content, posing new challenges for forgery detection.

\subsection{Video Detection Methods}
The rapid advancement of generative video models has led to the development of robust detection methods capable of identifying AI-generated content across diverse synthesis pipelines.

Existing approaches primarily focus on capturing spatial-temporal inconsistencies. DIVID \cite{turns} adopts a CNN-LSTM architecture to model frame-wise temporal dependencies. DeMamba \cite{demamba} leverages the Mamba architecture to enhance spatial-temporal representation learning. \cite{exposing} introduces a defect-based framework that addresses both local visual glitches and global temporal disruptions. DuB3D \cite{distinguish} proposes a dual-path architecture that combines spatial-temporal features with optical flow representations to improve robustness. AIGVDet \cite{aigvdet} explicitly separates the detection process into spatial and optical flow branches, allowing more targeted modeling of motion artifacts. \cite{matters} highlights the advantage of fusing appearance, motion, and geometric cues to enhance generalization across generation models.

In contrast to prior work that relies on optical flow maps or high-level temporal embeddings, our method explicitly captures motion inconsistencies through optical flow residuals, computed as differences between adjacent flow fields. This formulation offers a more sensitive and discriminative representation for detecting subtle temporal artifacts in AI-generated content.

\begin{figure*}[!t]
\centering
\includegraphics[width=0.98\linewidth]{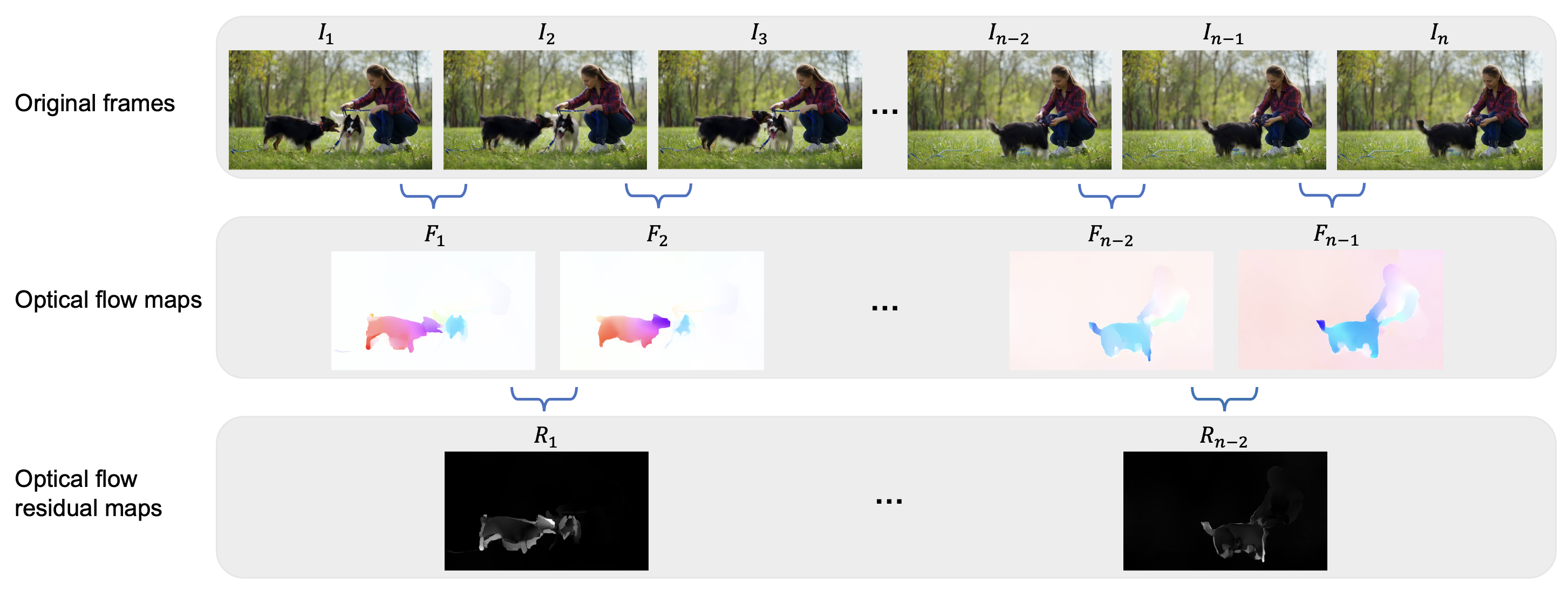}
\caption{\label{fig:preprocessing} Overview of the preprocessing pipeline.}
\end{figure*}

\section{Method}

\subsection{Motivation}
Existing video forgery detection methods often struggle to capture subtle temporal inconsistencies, particularly in diffusion-generated videos characterized by smooth and temporally coherent motion. Although some prior works incorporate optical flow to model temporal dynamics, such representations primarily capture first-order motion and emphasize global movement patterns. Consequently, they often fail to detect localized high-frequency motion anomalies introduced by generative models.

To address this limitation, we propose a second-order temporal representation based on optical flow residuals, computed as the difference between consecutive optical flow fields. This operation approximates the temporal derivative of motion, thereby amplifying fine-grained and abrupt changes that are typically imperceptible in optical flow maps. By attenuating consistent motion and highlighting local temporal deviations, optical flow residuals offer a more sensitive and discriminative representation to expose subtle artifacts in synthesized videos.

Motivated by this, we integrate optical flow residuals into a dedicated temporal branch within our detection framework, complementing appearance-level cues extracted from RGB frames.

\subsection{Overall Architecture}
Our detection framework adopts a dual-branch architecture to jointly capture spatial and temporal inconsistencies. As illustrated in Figure 1, the original branch takes RGB frames as input and focuses on visual artifacts, such as unnatural textures or object deformation. In parallel, the residual branch receives optical flow residuals and captures frame-to-frame motion irregularities.

Both branches employ the same architecture based on ResNet-50 \cite{resnet}, followed by classification heads that estimate the likelihood of manipulation. During inference, the outputs from both branches are fused to produce a final real or fake prediction.

\subsection{Preprocessing}
Before feeding the input video into the detection framework, a series of preprocessing steps are applied to prepare inputs for both spatial and temporal branches. The overall preprocessing pipeline is illustrated in Figure 2, which depicts the generation of RGB frames, optical flow maps, and optical flow residuals.

Given an input video $V \in \mathbb{R}^{H \times W \times C \times n}$, we first extract a sequence of RGB frames $\{ I_1, I_2, \dots , I_n \}$. These frames are fed into the original branch to detect appearance-based anomalies such as texture irregularities or boundary artifacts.

To capture temporal dynamics, we compute optical flow residuals, a more sensitive motion representation derived from the differences between consecutive optical flow maps.
We first estimate forward optical flow maps using RAFT \cite{raft}, a high-accuracy dense flow estimator that captures fine-grained motion while maintaining temporal coherence.

Specifically, forward optical flow maps between consecutive RGB frames are computed as follows:

\begin{equation}
F_t = \text{RAFT}(I_t, I_{t+1}), \quad t = 1, 2, \dots, n-1 \label{eq:01}
\end{equation}

Then, the optical flow residuals are obtained by:

\begin{equation}
R_t = F_{t+1} - F_t, \quad t = 1, 2, \dots, n-2 \label{eq:02}
\end{equation}

The resulting flow residuals $\{ R_1, R_2, \dots, R_{n-2} \}$ suppress globally consistent motion while amplifying localized temporal deviations. By highlighting frame-to-frame motion fluctuations, these residuals provide a more discriminative representation for identifying temporally implausible patterns in synthesized videos.

\subsection{Dual-Branch Detection Model}
The proposed detection framework comprises two structurally identical yet modality-specific branches. Both the original and residual branches employ a ResNet-50 backbone, followed by global average pooling and a fully connected classification head that estimates the probability of manipulation for each input.

The original branch processes a sequence of RGB frames $\{I_1, I_2, \dots, I_n\}$, aiming to detect spatial-level artifacts.
The residual branch takes as input the optical flow residuals $\{R_1, R_2, \dots, R_{n-2}\}$, which emphasize subtle temporal anomalies.

Each input, frame or residual map, is processed independently. The two branches are trained separately using binary cross-entropy loss, and their parameters are not shared, allowing each subnetwork to specialize in its respective modality.

At inference time, the final prediction score $P$ computed via a weighted fusion of the two branch outputs:

\begin{equation}
P = \alpha \cdot P_{\text{ori}} + \beta \cdot P_{\text{res}}, \quad   \alpha + \beta = 1 \label{eq:03}
\end{equation}

where $P_{\text{ori}}$ and $P_{\text{res}}$ represent the outputs of the original and residual branches, respectively. The fusion weights $\alpha$ and $\beta$ determine the relative contributions of spatial and temporal features. In our experiments, both hyperparameters are set to 0.5, thereby assigning equal importance to appearance and motion information. The final binary decision is obtained by thresholding the fused score at $T = 0.5$.

\section{Experiments}

\subsection{Datasets}
A diverse set of video datasets is utilized for training and evaluation, encompassing both text-to-video (T2V) and image-to-video (I2V) generation tasks. A detailed summary of the synthetic datasets used in our experiments is provided in Table 1. To ensure a fair comparison, the datasets marked as “Collection” in the “Source” column refer to datasets obtained from publicly available sources and are consistent with those used in previous studies \cite{wang, aigvdet}. In addition, the datasets labeled as “Generated”, Hunyuan and CogVideoX, were created using publicly released official code to further enrich the evaluation and demonstrate the generalizability of our method. These datasets cover a broad range of content, including human activities, natural scenes, and stylized environments.

\textbf{Training and Validation.}
The model is trained on 600 real and 600 synthetic videos, with an additional 250 real and 250 synthetic videos reserved for validation. The real videos are sourced from the YouTube-VOS dataset \cite{vos}, which contains diverse and naturally captured scenes. The synthetic training data are generated using two diffusion-based video generation models, Moonvalley and Sora. To ensure an unbiased evaluation of generalizability, all training and validation sets are kept disjoint from the test data.

\textbf{Testing.}
We evaluate the robustness and generalization ability of our method on synthetic videos generated by ten diffusion-based models. The real test videos are sampled from the GOT-10k dataset \cite{got}, which is disjoint from the training set and serves as an out-of-distribution benchmark. The synthetic test sets encompass both text-to-video (T2V) and image-to-video (I2V) tasks, and include outputs from a wide range of models such as Sora, Moonvalley, and Emu. This comprehensive benchmark enables a rigorous assessment of detection performance across various generative styles and previously unseen models.

\begin{table}
\centering
\caption{\label{tab:widgets}Details of the collected and generated synthetic video datasets used for evaluation.}
\begin{tabular}{llccc}
\hline
Task & Name & Count & Source \\
\hline
\multirow{11}{*}{T2V}
    & Emu \cite{emu} & 900 & Collection \\
    & Pika \cite{pika}  & 964 & Collection \\
    & Hotshot \cite{hotshot} & 500 & Collection \\
    & VideoCraft \cite{videocraft} & 1499 & Collection \\
    & NeverEnds \cite{neverends} & 1000 & Collection \\
    & VideoPoet \cite{videopoet} & 120 & Collection \\
    & Moonvalley \cite{moonvalley} & 3000 & Collection \\
    & Sora \cite{sora} & 48 & Collection \\
    & Hunyuan \cite{hunyuan} & 1000 & Generated \\
    & CogVideoX \cite{cogvideox} & 1000 & Generated \\
\hline
\multirow{3}{*}{I2V}
    & Pika \cite{pika} & 640 & Collection \\
    & NeverEnds \cite{neverends} & 642 & Collection \\
    & MoonValley \cite{moonvalley} & 642 & Collection \\
\hline
\end{tabular}
\end{table}

\begin{table*}[htbp]
\centering
\caption{Comparative results of baseline methods and the proposed model across diverse test datasets.}
\begin{tabular}{ll|cc|cc|ccc}
\toprule
\multirow{2}{*}{Task} & \multirow{2}{*}{Dataset} 
& \multicolumn{2}{c|}{Wang} 
& \multicolumn{2}{c|}{AIGVDet} 
& \multicolumn{3}{c}{Ours} \\
& & ACC & AUC & ACC & AUC & ACC & AUC & F1 \\
\midrule
\multirow{9}{*}{T2V} 
& Emu          & 97.3 & 99.7 & 94.1 & 99.3 & \textbf{100} & \textbf{100} & \textbf{100} \\
& Pika         & 68.9 & 89.5 & 89.5 & 95.5 & \textbf{99.6} & \textbf{99.9} & \textbf{99.6} \\
& Sora         & 55.2 & 89.0 & 82.3 & 92.1 & \textbf{94.8} & \textbf{99.5} & \textbf{94.5} \\
& Hotshot      & 56.0 & 81.5 & 93.3 & 98.0 & \textbf{99.9} & \textbf{100} & \textbf{99.9} \\
& NeverEnds    & 81.7 & 95.5 & 89.5 & 95.7 & \textbf{97.8} & \textbf{99.9} & \textbf{97.7} \\
& VideoPoet    & 76.7 & 95.6 & 93.8 & 97.7 & \textbf{94.2} & \textbf{99.9} & \textbf{93.8} \\
& VideoCraft   & 75.7 & 92.3 & 91.3 & 97.0 & \textbf{99.5} & \textbf{100} & \textbf{99.6} \\
& Moonvalley   & 99.3 & 100  & 95.1 & 100  & \textbf{100}  & \textbf{100} & \textbf{100} \\
& Hunyuan      &  -   &  -   &  -   &  -   & \textbf{99.6} & \textbf{100} & \textbf{99.6} \\
& CogVideoX     &  -   &  -   &  -   &  -   & \textbf{99.4} & \textbf{100} & \textbf{99.5} \\
\midrule
\multirow{3}{*}{I2V} 
& Pika         & 81.7 & 93.8 & 89.2 & 96.1 & \textbf{98.9} & \textbf{99.9} & \textbf{98.9} \\
& NeverEnds    & 74.8 & 92.5 & 91.0 & 96.6 & \textbf{98.8} & \textbf{99.9} & \textbf{98.8} \\
& Moonvalley   & 82.9 & 94.3 & 89.0 & 95.5 & \textbf{94.1} & \textbf{99.3} & \textbf{93.7} \\
\bottomrule
\end{tabular}
\label{tab:comparison}
\end{table*}

\subsection{Implementation Details}
The proposed method is implemented using the PyTorch framework and all experiments are conducted on an NVIDIA A6000 GPU. The architecture consists of two independent branches for spatial and temporal analysis, both employing ResNet-50 backbones initialized with ImageNet-pretrained weights. Each branch is trained separately using the Adam optimizer with an initial learning rate of 1e-4. Binary cross-entropy loss is used during training.  The learning rate is reduced by a factor of 10 after five consecutive epochs without improvement in validation accuracy, and training is terminated when the rate drops below 1e-6.

\subsection{Evaluation Metrics}
We use three evaluation metrics to assess the performance of our model: accuracy, AUC (area under the ROC curve) and the F1 score. These metrics provide a comprehensive evaluation by measuring the general accuracy of the model, its ability to distinguish between classes, and the balance between precision and recall.

\subsection{Experimental Results}

Table 2 summarizes the detection performance of our method in a diverse collection of synthetic video datasets in two generation tasks.

The proposed model demonstrates consistently strong performance across all benchmarks. In data sets such as Emu and Moonvalley, it achieves perfect scores of 100\% in accuracy, AUC, and F1 score, indicating high robustness under high-quality generation conditions. In addition, on datasets including Hotshot, VideoCraft, Pika, Hunyuan and CogVideoX, the model attains accuracy and F1 scores around 99.4\% to 99.9\%, with near-perfect AUC values, further confirming its reliability across varied generative patterns. Even on relatively challenging datasets such as Sora and VideoPoet, the model achieves accuracy values of 94.8\% and 94.2\%, respectively, with AUCs exceeding 99.3\%. These results underscore the model’s strong generalization ability across a wide range of generative styles and visual content domains.

Compared to existing methods, the proposed approach consistently outperforms both Wang \cite{wang} and AIGVDet \cite{aigvdet} across all evaluated datasets. For example, on the Sora dataset, our model achieves 94.8\% accuracy and 99.5\% AUC, representing improvements of 39.6 points in accuracy and 10.5 points in AUC over Wang, and 12.5 and 7.4 points, respectively, over AIGVDet. Similar performance gains are observed on other datasets such as Pika, NeverEnds, and VideoCraft, where the proposed method leads by considerable margins across all evaluation metrics.

Beyond overall performance, the model exhibits strong generalization across different generation modalities.  On the I2V datasets Pika and NeverEnds, it achieves 98.9\% accuracy and 99.9\% AUC, outperforming prior baselines by over 7 percentage points in accuracy. Furthermore, on the visually complex Moonvalley I2V dataset, the model attains 94.1\% accuracy and 99.3\% AUC, further demonstrating its robustness in detecting forgeries even under high-fidelity and structurally consistent synthesis conditions.

These results highlight the strength of our dual-branch design in capturing complementary spatial and temporal features, enabling superior generalization and adaptability across diverse generative videos.

\subsection{Comparison of Optical Flow Maps and Optical Flow Residuals}
To evaluate the effectiveness of optical flow residuals as a temporal representation, we conduct a comparative analysis against optical flow maps across various generated video datasets. As shown in Table 4, models that rely solely on the residual branch consistently outperform their optical flow counterparts in both accuracy and AUC, underscoring the superior discriminative capability of optical flow residuals as a temporal representation.

The performance advantage of optical flow residuals is particularly evident on challenging datasets. On the NeverEnds dataset, the model trained on residuals achieves 88.65\% accuracy and 98.73\% AUC, significantly outperforming the counterpart trained on optical flow maps, which achieves 65.95\% accuracy and 93.17\% AUC. A similar trend is observed on VideoCraft, where the residual-based model yields 86.39\% accuracy and 99.11\% AUC, compared to 68.61\% accuracy and 95.09\% AUC obtained by the flow-map-based model. On the Sora dataset, which contains highly realistic motion and complex temporal dynamics, the residual model achieves 79.17\% accuracy and 98.05\% AUC, while the model using optical flow maps reaches only 60.42\% accuracy and 96.05\% AUC.

These findings suggest that optical flow residuals offer a more sensitive and informative temporal representation by capturing frame-to-frame variations in motion. While optical flow maps primarily encode first-order motion, flow residuals approximate second-order dynamics, making them more effective at revealing subtle temporal inconsistencies introduced during generation. This enhanced sensitivity provides a stronger supervisory signal for AI-generated video forgery detection.

\begin{table}[htbp]
\centering
\caption{Comparative evaluation of optical flow maps and optical flow residuals.}
\label{tab:widgets}
\begin{tabular}{ll|cc|cc}
\toprule
\multirow{2}{*}{Task} & \multirow{2}{*}{Dataset} 
& \multicolumn{2}{c|}{Optical Flow} 
& \multicolumn{2}{c}{Flow Residual} \\
& & ACC & AUC & ACC & AUC \\
\midrule
\multirow{9}{*}{T2V} 
& Emu         & 99.28 & 99.88 & 99.48 & 99.95 \\
& Pika        & 65.51 & 95.95 & 69.43 & 96.72 \\
& Sora        & 60.42 & 96.05 & 79.17 & 98.05 \\
& Hotshot     & 91.90 & 97.82 & 96.00 & 99.28 \\
& NeverEnds   & 65.95 & 93.17 & 88.65 & 98.73 \\
& VideoPoet   & 63.75 & 93.19 & 91.25 & 99.08 \\
& VideoCraft  & 68.61 & 95.09 & 86.39 & 99.11 \\
& Moonvalley  & 79.38 & 98.99 & 98.87 & 99.94 \\
& Hunyuan     & 70.25 & 98.02 & 79.60 & 98.54 \\
& CogVideoX   & 73.20 & 97.31 & 83.00 & 98.07 \\
\midrule
\multirow{3}{*}{I2V} 
& Pika        & 69.30 & 96.29 & 80.55 & 96.26 \\
& NeverEnds   & 74.92 & 96.41 & 90.03 & 98.04 \\
& Moonvalley  & 65.97 & 94.80 & 84.42 & 96.68 \\
\bottomrule
\end{tabular}
\end{table}

\section{Conclusion}
In this work, we propose a dual-branch framework for video forgery detection that jointly models spatial and temporal inconsistencies. By leveraging RGB frames and optical flow residuals, the proposed method captures both appearance-level artifacts and fine-grained motion irregularities in AI-generated videos.
Extensive experiments conducted on a diverse set of datasets demonstrate that our approach consistently outperforms existing state-of-the-art methods. In particular, it exhibits strong generalization capability on challenging diffusion-generated content across both text-to-video and image-to-video tasks. These results highlight the discriminative strength of optical flow residuals as a temporal representation and validate the effectiveness of the proposed architecture in achieving robust and accurate detection of synthetic videos.

Nevertheless, we observe that the residual branch shows room for improvement on highly realistic videos, especially those generated by advanced models such as Sora, Pika, and Hunyuan. Future work may explore more adaptive temporal modeling techniques to address these challenging cases.




%



\bibliographystyle{IEEEtran}

\end{document}